\documentclass[letterpaper, 10 pt, conference]{ieeeconf}  

\IEEEoverridecommandlockouts                              

\usepackage{multirow,graphicx}

\usepackage{booktabs}

\usepackage{tabularx}
\usepackage{adjustbox}
\usepackage[table,xcdraw]{xcolor}
\usepackage{multirow}
\usepackage{rotating}
\usepackage{pdflscape}
\usepackage{makecell}
\usepackage{bm}
\usepackage{empheq}
\usepackage{amsfonts}
\usepackage{hyperref}

\usepackage{pgfplots}
\usepackage{algorithm, algorithmicx, algpseudocode}
\usepackage{pgfplots}
\pgfplotsset{compat=newest}
\usetikzlibrary{plotmarks}
\usetikzlibrary{arrows.meta}
\usepgfplotslibrary{patchplots}
\usepackage{grffile}
\usepackage{amsmath}
\usepackage{listings}

\usepackage{siunitx}
\usepackage{float}

\usepackage{etoolbox}
\makeatletter
\patchcmd{\@makecaption}
{\scshape}
{}
{}
{}
\makeatother

\title{\LARGE \bf
	Disturbance Estimation and Rejection for High-Precision Multirotor Position Control
}

\author{Daniel Hentzen$^{1,2}$, Thomas Stastny$^{2}$, Roland Siegwart$^{2}$, Roland Brockers$^{1}$
\thanks{$^{1}$Authors are with the Jet Propulsion Laboratory, California Institute of Technology, 4800 Oak Grove Dr, Pasadena, CA 91109, USA
	{\tt\small hentzend@gmail.com}; {\tt\small brockers@jpl.nasa.gov}}%
\thanks{$^{2}$Authors are with the Swiss Federal Institute of Technology (ETH)
Zurich, Autonomous Systems Lab, Leonhardstrasse 21,
8092 Zurich, Switzerland
        {\tt\small (firstname.lastname)@ethz.ch}}%
}

\newcommand{\Assert}[1]{%
  \State \textbf{assert} #1%
}

\makeatletter
\newcommand{\algorithmfootnote}[2][\footnotesize]{%
  \let\old@algocf@finish\@algocf@finish
  \def\@algocf@finish{\old@algocf@finish
    \leavevmode\rlap{\begin{minipage}{\linewidth}
    #1#2
    \end{minipage}}%
  }%
}
\makeatother

\algrenewcommand{\alglinenumber}[1]{\color{gray}\footnotesize#1:}

\algnewcommand\algorithmicswitch{\textbf{switch}}
\algnewcommand\algorithmiccase{\textbf{case}}
\algnewcommand\algorithmicassert{\texttt{assert}}
\algnewcommand\Assert[1]{\State \algorithmicassert(#1)}%
\algdef{SE}[SWITCH]{Switch}{EndSwitch}[1]{\algorithmicswitch\ #1\ \algorithmicdo}{\algorithmicend\ \algorithmicswitch}%
\algdef{SE}[CASE]{Case}{EndCase}[1]{\algorithmiccase\ #1}{\algorithmicend\ \algorithmiccase}%
\algtext*{EndCase}%

\definecolor{mygreen}{rgb}{0,0.6,0}
\definecolor{mygray}{rgb}{0.5,0.5,0.5}
\definecolor{mymauve}{rgb}{0.58,0,0.82}
\lstdefinestyle{BashInputStyle}{
  language=bash,
  basicstyle=\small\sffamily,
  numbers=left,
  numberstyle=\tiny,
  numbersep=3pt,
  frame=tb,
  columns=fullflexible,
  backgroundcolor=\color{yellow!20},
  moredelim=**[is][\color{gray}]{@}{@},
  belowskip=-1.2 \baselineskip
}

\begin{document}

\maketitle
\thispagestyle{empty}
\pagestyle{empty}

\begin{abstract}
Many multirotor Unmanned Aerial Systems applications have a critical need for precise position control in environments with strong dynamic external disturbances such as wind gusts or ground and wall effects. Moreover, to maximize flight time, small multirotor platforms have to operate within strict constraints on payload and thus computational performance. In this paper, we present the design and experimental comparison of Model Predictive and PID multirotor position controllers augmented with a disturbance estimator to reject strong wind gusts up to 12 m/s and ground effect. For disturbance estimation, we compare Extended and Unscented Kalman filtering. In extensive in- and outdoor flight tests, we evaluate the suitability of the developed control and estimation algorithms to run on a computationally constrained platform. This allows to draw a conclusion on whether potential performance improvements justify the increased computational complexity of MPC for multirotor position control and UKF for disturbance estimation.
\end{abstract}

\section*{Supplementary Material}
Video: \url{https://youtu.be/-1PvZ5YBIuw}{}

\section{INTRODUCTION}
\subsection{Motivation}
For a majority of proposed multirotor Unmanned Aerial Systems (UAS) applications, successful deployment relies on the ability of the flight controller to handle unpredictable and potentially adverse conditions inherent to real-world outdoor environments. In particular, the control system must accurately track position references despite the presence of strong and dynamic disturbances, such as wind gusts or ground and wall effects. Examples of relevant applications include outdoor docking maneuvers, for instance for autonomous recharging \cite{Brommer2018}, industrial inspection \cite{Omari2015} or aerial package delivery \cite{DAndrea2014}. In the latter application for instance, the vehicle must follow a path and perform landings accurately despite the presence of a static disturbance (due to the varying payload mass) as well as wind gusts and ground effects that occur during flight through urban landscapes.

Recent work has shown that the problem of accurate multirotor position control is well suited to a Model Predictive Control (MPC) formulation, since it is able to compute optimized control inputs while accounting for state and input constraints \cite{Kamel2017}. However, the capability of multirotor MPC to deal with strong dynamic disturbances remains uncertain. While previous work has shown that estimates of external disturbances can be computed online and used within the flight control system to improve control performance (see Section \ref*{subsec:related-work}), the suitability of this approach with model predictive position controllers is untested in practice and shall thus be studied in this work.

Moreover, we aim to gather conclusive experimental insights on whethe the choice of MPC over classical control formulations for multirotor position control in real-world environments is justified by substantial performance gains, considering MPC's higher complexity and computational cost.


\subsection{Contribution}
\begin{figure}
	\centering
	\includegraphics[width=1.0\linewidth]{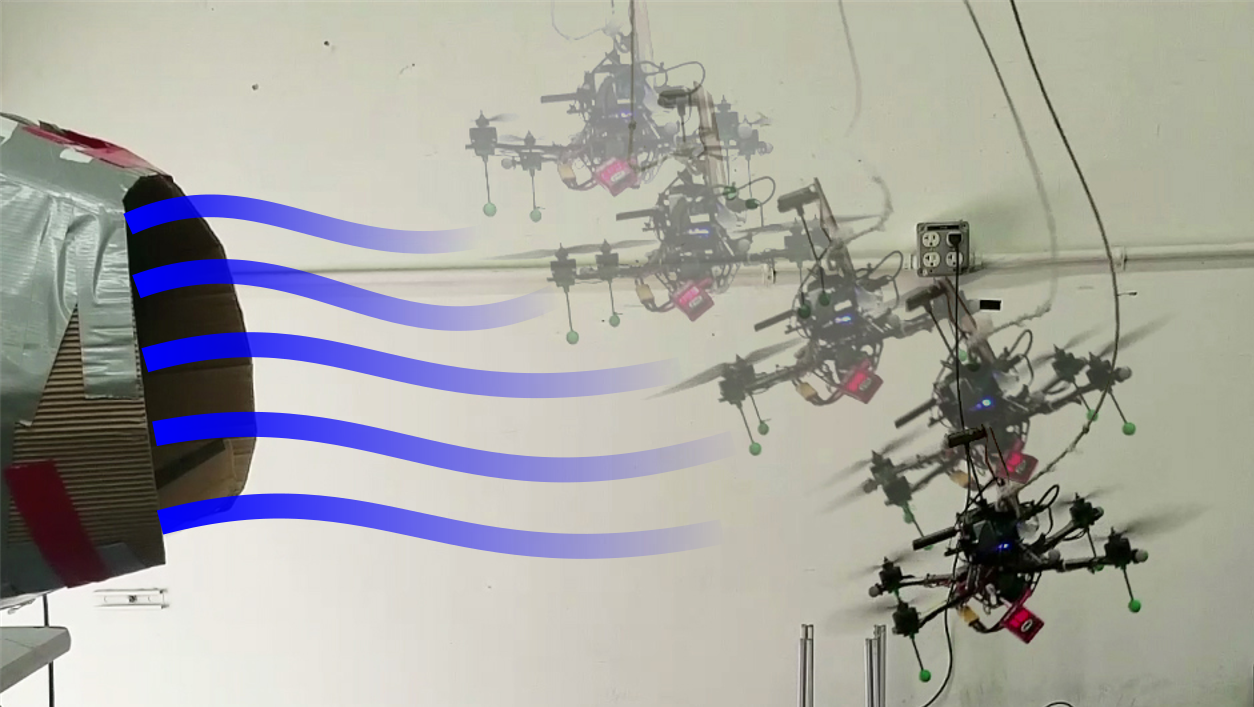}
	\caption{Example of the effect of a strong wind gust on a quadrotor.}
	\label{fig:setupgust}
\end{figure}
In this work, we address the above research questions by presenting a first-of-a-kind experimental comparison of two position controllers for a small quadrotor platform: a Nonlinear Model Predictive controller and a conventional PID controller. Both control formulations are augmented with a disturbance estimation module that enables the rejection of dynamic disturbances. For disturbance estimation, we compare Extended and Unscented estimators. In extensive flight tests, we assess the disturbance estimation and rejection performance of the presented algorithms in the presence of a) ground effect and b) wind gusts up to 12 m/s. We also measure the computational footprint of the control and estimation algorithms on a small microprocessor. The disturbance estimator comparison is first performed in a Vicon motion capture environment with near-perfect state estimation (as in existing work on multirotor disturbance estimation), but also repeated for the first time outdoors with degraded GPS-based state estimation. We conclude that the increased complexity of MPC is justified by significant improvements in position tracking under strong disturbances compared to PID control. Moreover, we determine that Unscented filtering does not offer compelling performance gains over Extended filtering to justify its higher computational cost.

%

\subsection{Related Work}
\label{subsec:related-work}
Robust control formulations for multirotor applications, such as $\mathcal{H}_\infty$ \cite{Raffo2010a} or Robust MPC \cite{Alexis2016a}, have been developed with the aim to provide robust performance guarantees given an unknown but bounded disturbance. However, in unpredictable real-world environments, the disturbance bounds have to be chosen conservatively to guard against worst-case disturbances, resulting in degraded control performance. Alternatively, adaptive control formulations are designed to determine uncertain system parameters online. The authors in \cite{Achtelik2011a} present quadrotor controllers based on Model Reference Adaptive Control and show improved reaction to perturbations such as a sudden loss-of-thrust or weight disturbances compared to non-adaptive baseline controllers. In \cite{Pereida2017}, an $\mathcal{L}_1$ adaptive controller is employed to reject disturbances caused by a suspended weight pendulum. A known drawback of adaptive controllers is a lack of robustness due to phenomena such as bursting \cite{Achtelik2011a}. This motivates the need for robustness modifications that limit the adaptation performance and thus the rejection of dynamic disturbances.

\begin{figure}[t]
	\vspace{5mm}
	\centering
	\includegraphics[width=0.45\linewidth]{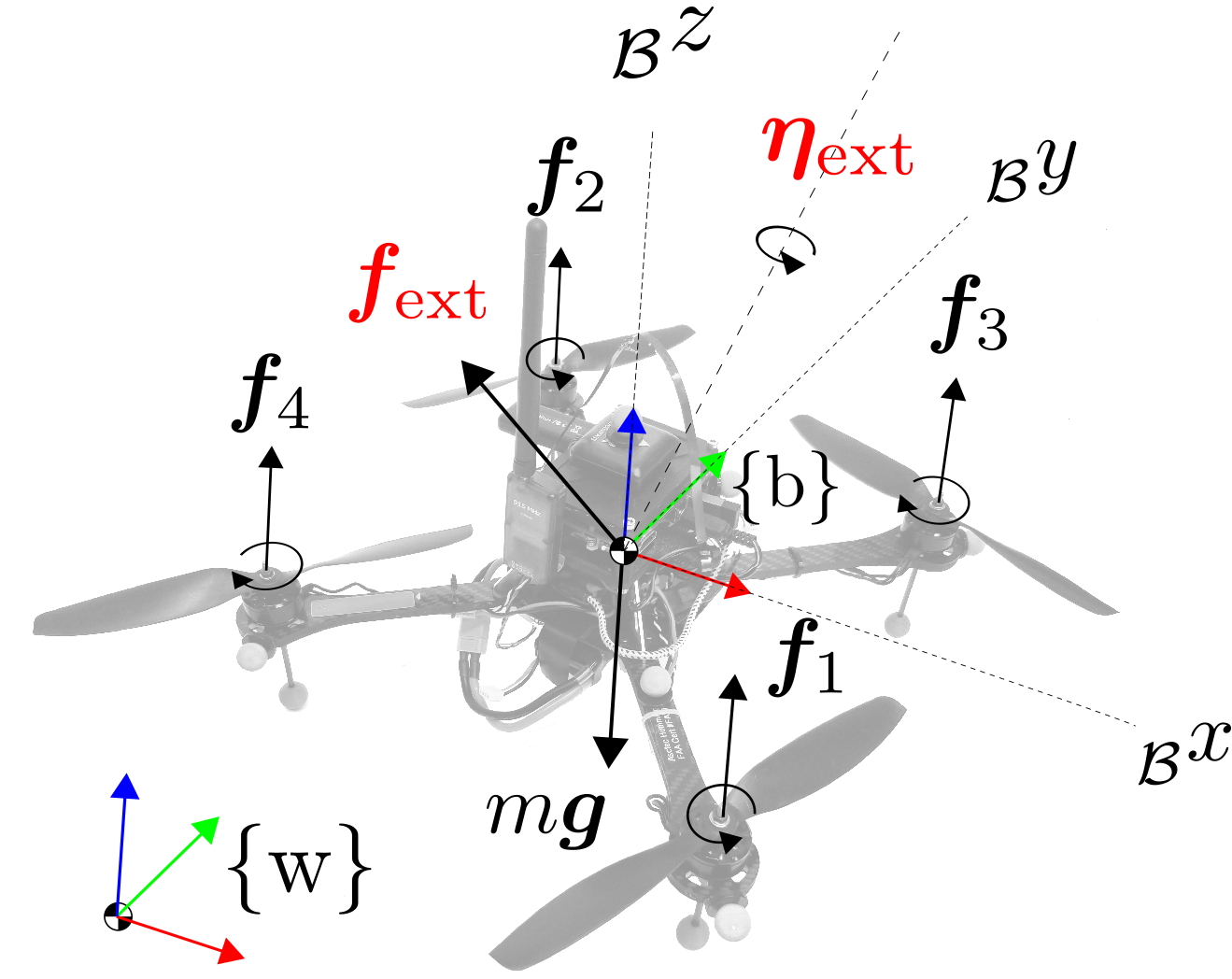}
	\caption{Quadrotor free-body diagram and coordinate systems.}
	\label{fig:coordinate_systems}
\end{figure}

A promising approach to reject dynamic disturbances is to observe the disturbance online and compute a compensating control input. In \cite{Ruggiero2014}, a momentum-based estimator of external forces is used in conjunction with an impedance controller. In \cite{Smeur2018}, a Nonlinear Inversion-based position controller uses filtered accelerometer measurements to reconstruct and compensate the acceleration caused by external disturbance, resulting in significant improvement of disturbance rejection. In \cite{Tomic2014}, \cite{Johansson2016}, the external force \cite{Tomic2014} \cite{Johansson2016} and torque vectors \cite{Tomic2014} are estimated using acceleration and velocity measurements.  However, the quality of the above disturbance estimation methods degrades with sensor noise, as the proposed observers are deterministic and do not account for measurement or process noise models \cite{McKinnon2016}. It is thus preferrable to use stochastic estimators if sufficient computational resources are available. In \cite{Tagliabue2017, McKinnon2016}, two disturbance force and torque estimators, based on Unscented Kalman Filtering are presented. Yet it is unclear whether the use of the more computationally costly UKF over an EKF is justified. In \cite{Kamel2017}, external disturbance forces are estimated by an EKF. However, no experimental results for the disturbance estimator are presented. The disturbance estimates are used within a nonlinear MPC framework to achieve accurate trajectory tracking in the presence of a 11 m/s wind disturbance. In this work, we use the same nonlinear MPC approach as in \cite{Kamel2017} in a path following formulation, extend it with optional soft constraints and compare its performance to that of an augmented PID controller.

\section{SYSTEM MODELING}

\subsection{Vehicle Dynamics}
\label{subsec:vehicle_dynamics}

We consider a quadrotor controlled by four individual rotor thrusts of magnitude $f_i$ and rotor drag torques of magnitude $\eta_i$, yielding a collective thrust $T$ and body torque $\bm{\eta}_{\text{prop}}$. Additionally, an external disturbance force $\bm{f}_{\text{ext}}$ and external disturbance torque $\prescript{}{\mathcal{B}}{\bm{\eta}}_{\text{ext}}$ act on the vehicle. Figure \ref{fig:coordinate_systems} shows the free-body diagram and coordinate systems.

%

We can derive the translation dynamics of the quadrotor in world frame using Newton's $2^{\text{nd}}$ law for rigid bodies: 
\begin{empheq}[]{align}
\label{eq:translation_dyn_1}
\dot{\bm{p}} &= \bm{v}, \\
\label{eq:simple_velocity_dynamics_1}
\dot{\bm{v}} &=
\frac{1}{m} \bigl(\mathbf{C}_{\mathcal{B}}^{\mathcal{W}} T \bm{e}_z^b + \bm{f}_{\text{ext}} \bigr) + \bm{g},
\end{empheq}
where $\mathbf{C}_{\mathcal{B}}^{\mathcal{W}}$ is the rotation matrix from body to world frame.

The quadrotor attitude dynamics are derived using the Euler equation for rigid bodies and quaternion rate of change kinematics:
\begin{empheq}[]{align}
\nonumber
\dot{\bm{q}} &= \frac{1}{2}
\begin{bmatrix}
0 \\ \bm{\omega}
\end{bmatrix}\otimes\bm{q},
\\
\nonumber
\label{eq:simple_omega_dynamics_1}
\prescript{}{\mathcal{B}}{\dot{\bm{\omega}}} &= J^{-1}\Biggl(
\prescript{}{\mathcal{B}}{\bm{\eta}_{\text{prop}}} + \prescript{}{\mathcal{B}}{\bm{\eta}_{\text{ext}}} -\prescript{}{\mathcal{B}}{\bm{\omega}}\times(J\prescript{}{\mathcal{B}}{\bm{\omega}}) \Biggr),
\end{empheq}
where $\otimes$ represents a quaternion multiplication.

\subsection{Thrust and Drag Torque Maps}
\label{subsec:thrust_torque_maps}

\begin{figure}[t]
	\vspace{5mm}
	\centering
	\includegraphics[width=0.55\linewidth]{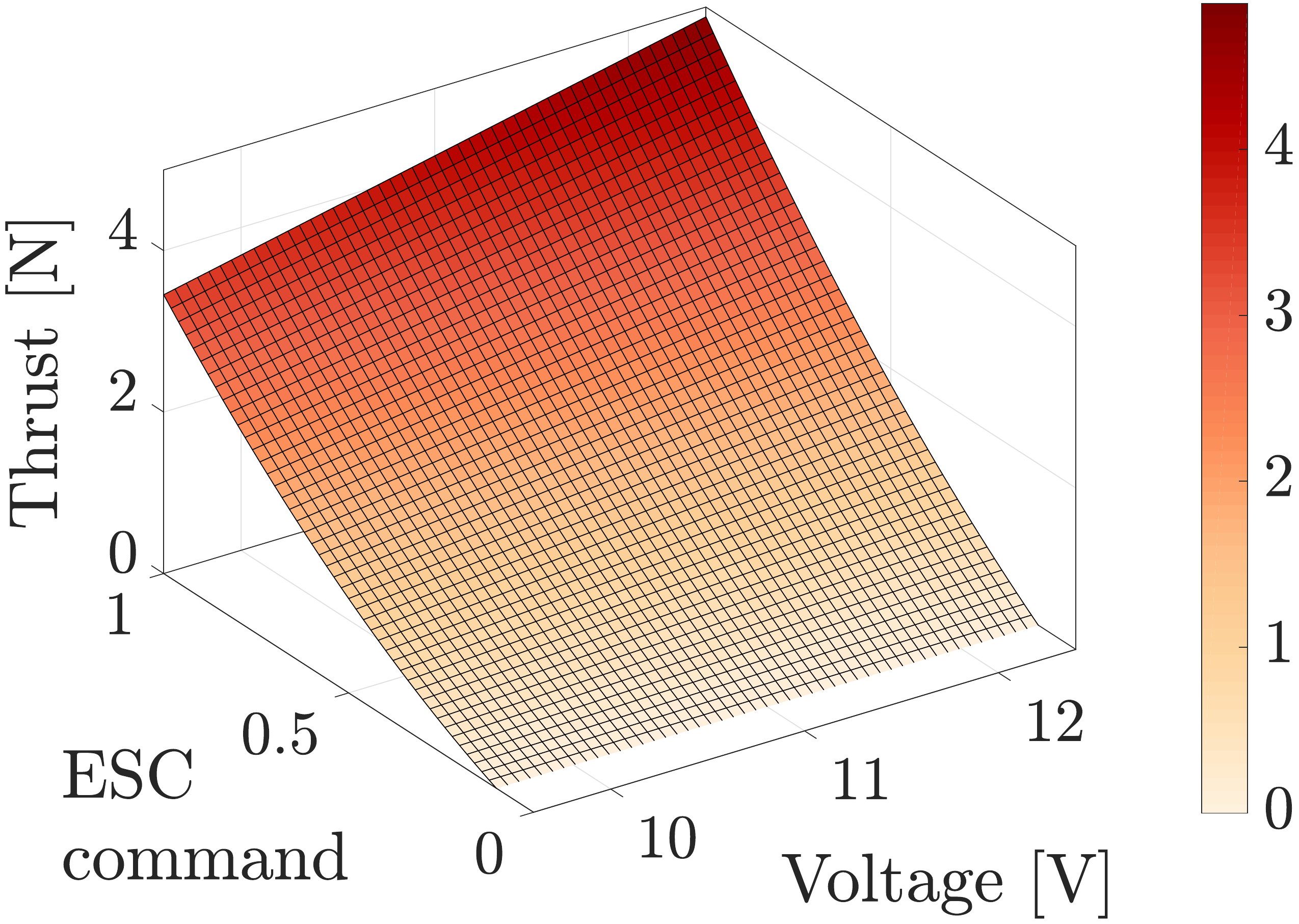}
	\caption{Identified single rotor thrust map for different ESC commands (here shown normalized for a PWM range of 1000 - 2000) and different battery voltage levels.}
	\label{fig:thrustmap}
\end{figure}

The disturbance estimation and control algorithms presented below require knowledge and control of the output rotor thrusts and drag torques. Due to the lack of both closed-loop motor control and motor RPM measurements on our platform, we identify a map from the commanded ESC commands $u_{\text{ESC},i}$ to rotor thrust and torque magnitudes $f_i$ and $\eta_i$ respectively. 
We use the same quadratic map between ESC command and output thrust/torque as in \cite{Faessler2017} and extend it with a linear dependency on battery voltage. Figure \ref{fig:thrustmap} shows an example of the identified voltage-dependent thrust map for a single rotor.

\section{DISTURBANCE ESTIMATION}
\label{sec:disturbance_estimation}
To estimate the external force $\bm{f}_{\text{ext}}$ and torque $\bm{\eta}_{\text{ext}}$ acting on the quadrotor, we design and implement two different disturbance estimators: an Extended and an Unscented Kalman Filter.

\begin{figure}
	\vspace{4mm}
	\includegraphics[width=1.00\linewidth]{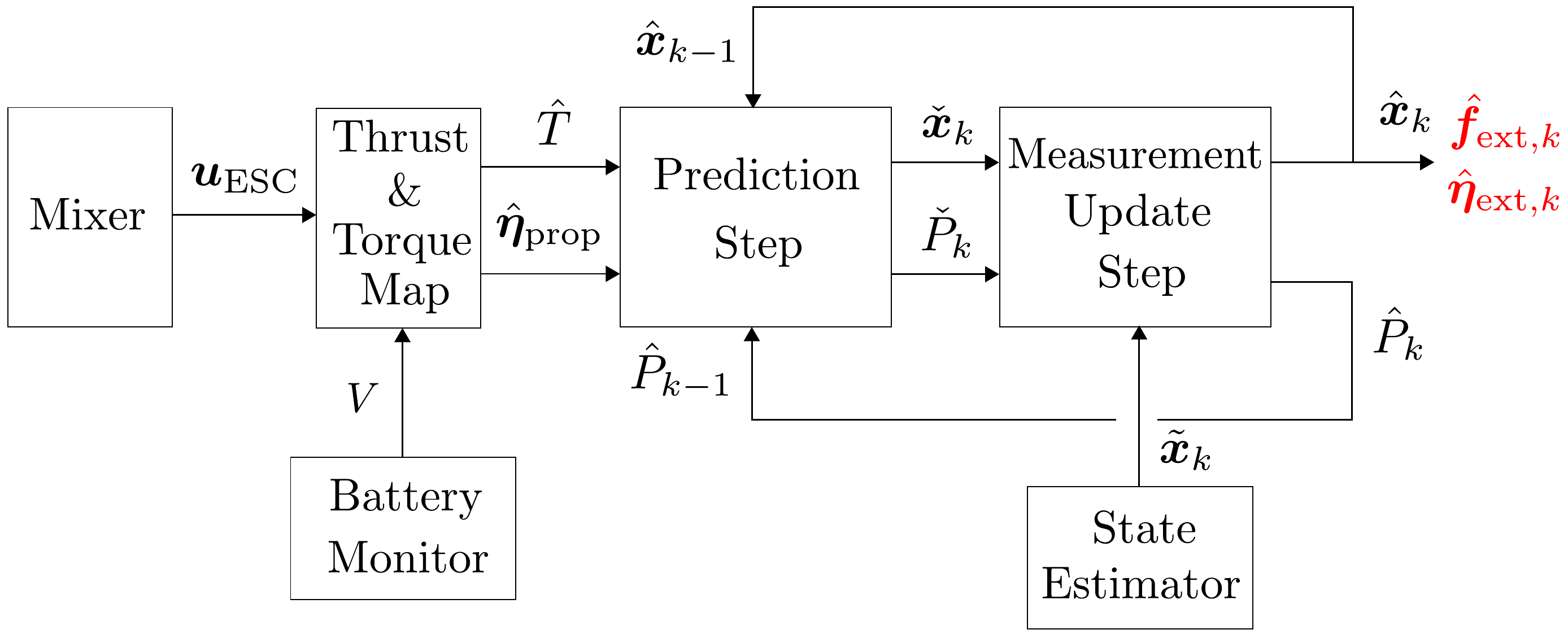}
	\caption{Recursive filtering architecture for disturbance estimation.}
	\label{fig:disturbanceestimationarchitecture}
\end{figure}

Due to the shared recursive nature, we employ an identical architecture for both filters, illustrated in Figure \ref{fig:disturbanceestimationarchitecture}. The individual ESC motor commands $\bm{u}_{\text{ESC}}$ are converted into an estimated net thrust $\hat{T}$ and body torque $\hat{\bm{\eta}}_{\text{prop}}$ using our thrust and drag torque calibration (Section \ref{subsec:thrust_torque_maps}). The prediction step then propagates the vehicle dynamics to yield a predicted state mean and covariance. The vehicle state includes the disturbance terms to be estimated. The predicted state mean and covariance are corrected using measurements provided by the onboard state estimator, yielding the estimated disturbance force/torque mean and covariance. 

\subsection{Extended Kalman Filter}

\paragraph{System State}
We define the system state $\bm{x}$ as the vehicle position, velocity, attitude and angular rates in body frame and augment it with the external forces and torques:
\[\bm{x} := \bigl[\bm{p}^T \quad \bm{v}^T \quad \phi \quad \theta \quad \psi \quad \prescript{}{\mathcal{B}}{\bm{\omega}}^T \quad \bm{f}_{\text{ext}}^T \quad \bm{\eta}_{\text{ext}}^T\bigr]^T.\]
Despite the gimbal-lock singularity at $\theta = \pi/2$, we choose an Euler attitude representation for the EKF design. In contrast to the quaternion representation, the linearization is simplified and does not require quaternion normalization checking \cite{Barfoot}.

The evolution of the system state $\bm{x}$ is subject to zero-mean Gaussian process noise $\bm{v}$ with covariance $Q$.

%

\paragraph{System Input}
The system input $\bm{u}$ consists of the estimated collective propeller thrust $\hat{T}$ and body torques $\prescript{}{\mathcal{B}}{\hat{\bm{\eta}}}_{\text{prop}}$:
\[\bm{u} := \bigl[\hat{T} \quad \prescript{}{\mathcal{B}}{\hat{\bm{\eta}}}_{\text{prop}}^T\bigr]^T.\]

\paragraph{Measurement Model}
The measurement vector $\bm{z} = \tilde{\bm{x}}$ is provided by the state estimator:
\[\bm{z} := \bigl[\tilde{\bm{p}}^T \quad \tilde{\mathbf{v}}^T \quad \tilde{\phi} \quad \tilde{\theta} \quad \tilde{\psi} \quad \tilde{\bm{\omega}}^T\bigr]^T.\]
The measurements are subject to additive zero-mean Gaussian process noise $\mathbf{w}$ with covariance $R$.


\paragraph{Discrete System Dynamics}
To formulate the EKF prediction step for external force and torque estimation, we discretize the quadrotor dynamics presented in Section \ref{subsec:vehicle_dynamics} (with Euler angles replacing quaternions) using a forward Euler scheme.
To allow a flexible application of the disturbance estimator, we model the force and torque dynamics as a Gaussian random walk:

\begin{empheq}[]{align}
\nonumber
&\bm{f}_{\text{ext},k} = \bm{f}_{\text{ext},k-1} + \mathbf{v}_f,\\
\nonumber
&\prescript{}{\mathcal{B}}{\bm{\eta}}_{\text{ext},k} = \prescript{}{\mathcal{B}}{\bm{\eta}}_{\text{ext},k-1} + \mathbf{v}_\eta.
\end{empheq}
The process noise value $\mathbf{v}$ is a tuning parameter that allows to control the convergence speed of the estimate, at the cost of a noisier output signal.


\paragraph{Prediction and Measurement Update Steps}
The prediction and measurement update steps are performed using the standard EKF equations \cite{Simon2006}.

\subsection{Unscented Kalman Filter}

In the UKF, we use the same state as in the EKF, but replace the Euler angle representation with a quaternion representation. Since there is no need to derive a Jacobian matrix, the quaternion representation is easy to handle in the UKF formulation and avoids the gimbal-lock singularity at $\theta=\pi/2$. For the attitude representation, we follow the same multiplicative approach used in \cite{Tagliabue2017}. The mean attitude is represented using singularity-free, unit-norm quaternions, whereas the rotational uncertainty is parameterized and passed through the Unscented Transform in the form of Modified Rodrigues Parameters (MRP). 

\paragraph{State, Input, Measurements} 
We maintain the same state, input and measurement vectors as in the EKF, but replace Euler angles with quaternions. The process covariance matrix $Q$, the noise covariance matrix $R$ and the measurement model matrix $H$ follow their EKF counterparts.

\paragraph{Discrete System Dynamics} The UKF uses the same discrete translation dynamics as the EKF, while the discrete quaternion-based rotation dynamics are detailed in \cite{Crassidis2003}.

%

\paragraph{Prediction and Measurement Update Steps} The prediction and measurement update steps are performed using the standard UKF equations \cite{Simon2006, Uhlmann1995}. For switching between the full attitude quaternion and the MRPs we follow the approach outlined in \cite{Tagliabue2017}. 

\section{POSITION CONTROL WITH DISTURBANCE REJECTION}

\subsection{Control System Overview}
\begin{figure}
	\vspace{4mm}
	\centering
	\includegraphics[width=1.0\linewidth]{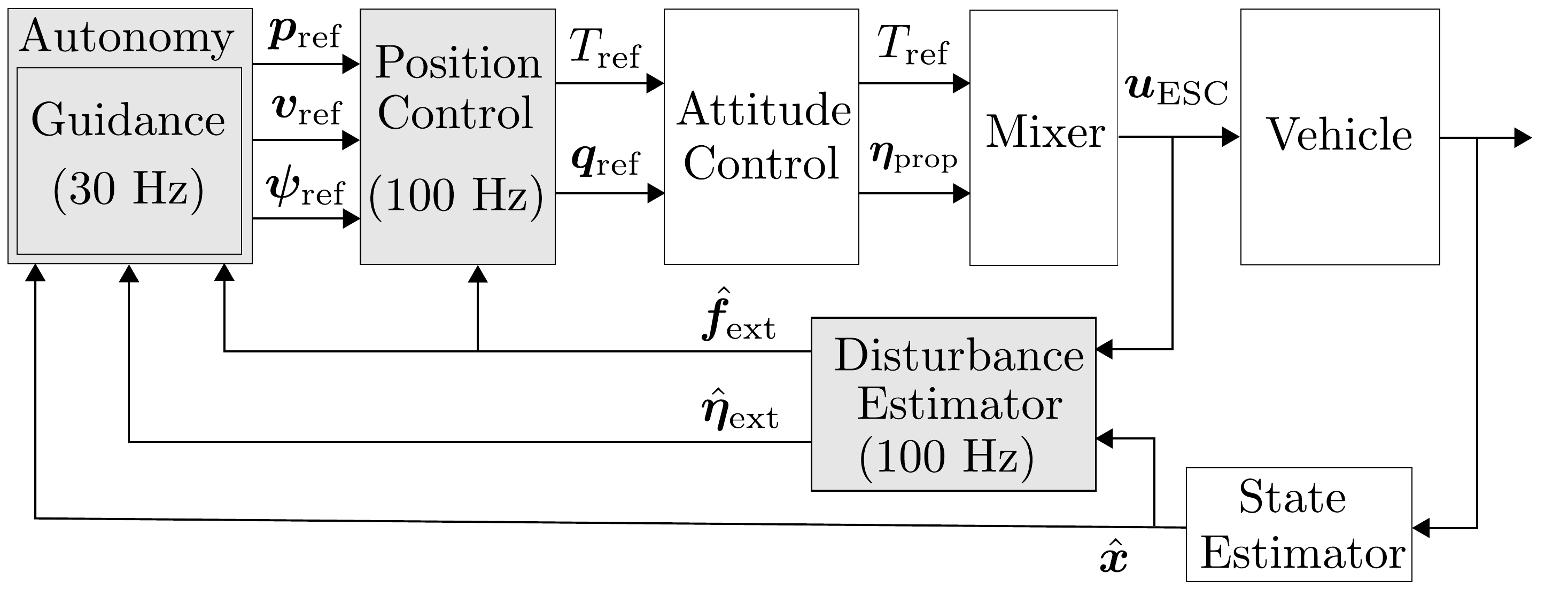}
	\caption{Guidance and control system architecture. Shaded modules are executed on an \textit{Odroid XU4} companion computer, while white modules are running on a \textit{Pixhawk mini} microcontroller autopilot.}
	\label{fig:controlarchitecture}
\end{figure}

Figure \ref{fig:controlarchitecture} provides a schematic overview of the entire flight control system and how it interfaces with the guidance and disturbance estimation modules described in Sections \ref{subsec:guidance} and \ref{sec:disturbance_estimation}. The control system is based on a cascaded architecture. Given a state estimate as well as position and velocity references issued by the guidance module, the position controller computes a desired acceleration vector. The latter is then transformed into a collective thrust and a reference attitude quaternion, which are passed as references to the low-level attitude controller. Both of the presented position controllers use the force estimates provided by the disturbance estimator to reject the external disturbance acting on the vehicle. While the torque estimates could be used within the attitude controller, we decide to limit the scope of the active disturbance rejection to the position controller, in order to assess performance in combination with the widely used off-the-shelf \textit{PX4} attitude controller.

\subsection{Path Following Guidance}
\label{subsec:guidance}
To maximize robustness to large external disturbances, we choose to implement a path following guidance system (as opposed to trajectory tracking), which guides the vehicle to the closest point on the spatially-parameterized reference path while maintaining a pre-defined reference velocity along the path tangent. In the presence of large disturbances, this prevents an accumulation of the position and velocity errors over time and thus excessively aggressive flight maneuvers without the need for computationally expensive motion replanning. Such an approach is well suited for applications that need to achieve a certain level of precision in order to avoid hazards, such as precision landing \cite{Brommer2018} for instance, where a precise approach and a safe touch-down (for example on a recharging station) is critical.

\subsection{PID Control Formulation}
The PID controller computes the desired acceleration components as follows:
\begin{equation}
a_{\text{des},i} = k_{p,i} {e}_{p,i} + k_{d} e_{v,i} + k_{i} \int {e}_{p,i} - \frac{1}{m} \hat{f}_{\text{ext},i},
\end{equation}
where $i \in \{x,y,z\}$.
The conventional PID structure is augmented with a feedforward disturbance acceleration term to compensate for the estimated force disturbance $\bm{\hat{f}}_{\text{ext}}$. The output acceleration vector $\bm{a}_{\text{des}}$ is then transformed into a reference attitude quaternion and collective thrust \cite{Brescianini2013abc}.

\subsection{MPC Control Formulation}

\paragraph{System Model}
We implement a nonlinear MPC based on the full nonlinear system dynamics. In \cite{Kamel2017} the authors showed that nonlinear MPC offers superior performance to linear MPC for multirotor trajectory tracking, while fast loop times can be achieved for both formulations thanks to recent advances in open-source nonlinear solvers \cite{Houska2011a}.

The translation dynamics are formulated in (\ref{eq:translation_dyn_1}) and (\ref{eq:simple_velocity_dynamics_1}). The model includes the external disturbance force acting on the vehicle. All unmodeled dynamics (such as aerodynamic drag) are lumped together in this disturbance force. We assume a constant disturbance force over the MPC prediction horizon.
As in \cite{Kamel2017}, we identify the attitude dynamics as a first-order system. This allows us to deploy the MPC position controller with arbitrary (open- or closed-source) attitude controllers. 


\paragraph{Optimal Control Problem}
The system state is given by the vehicle position, velocity and Euler angles:
\[
\bm{x} = \bigl[\bm{p}^T \quad \bm{v}^T \quad \phi \quad \theta \quad \psi \bigr]^T.
\]
The control input to be computed consists of the net thrust and a reduced roll-pitch attitude in the form of Euler angles:
\[
\bm{u} = \bigl[T_{\text{ref}} \quad \phi_{\text{ref}} \quad \theta_{\text{ref}}\bigr]^T
\]
The control inputs serve as the reference to the low-level attitude controller. The yaw reference is directly passed to the attitude controller.

We can now formulate the continuous, nonlinear optimal control problem (OCP) to be solved by the MPC at each iteration:
\begin{equation}
\label{eq:ocp}
\begin{split}
\min_{\bm{U}} &\int_{t=0}^{T} \Bigl[\bm{\bar{x}}(t)^T Q_x \bm{\bar{x}}(t)
+ \bm{\bar{u}}(t)^T
R_u \bm{\bar{u}}(t) \Bigr] dt\\
&+ \bm{\bar{x}}(T)^T P_N \bm{\bar{x}}(T),\\
\text{subj. to} \quad &\dot{\bm{x}} = {{f(\bm{x},\bm{u})}}, \quad \text{(system dynamics)}\\
&\bm{u}(t) \in \mathbb{U}, \quad \text{(thrust, roll, pitch input constraints)}\\
&\bm{x}(0) = \bm{x}(t_0), \quad \text{(initial condition)}
\end{split}
\end{equation}
where $\bm{\bar{x}}$ and $\bm{\bar{u}}$ denote the state and input error.
The references for the states and inputs are provided by the guidance module.
%
%
%
%
The constrained input space is given by
\[
\left.
\mathbb{U} = \Biggl\{\bm{u} \in \mathbb{R}^3 \right\vert \begin{bsmallmatrix}
T_{\text{min}} \\ -\phi_{\text{max}} \\-\theta_{\text{max}} \end{bsmallmatrix} 
\leq \bm{u} \leq \begin{bsmallmatrix}
T_{\text{max}} \\ \phi_{\text{max}} \\ \theta_{\text{max}}
\end{bsmallmatrix}\Biggr\}.
\]

\paragraph{Soft-Constrained MPC}
We implement optional constraints on the cross-error along the x and y axes. To reduce the risk of infeasibility, we  use a soft constraint, or slack constraint. The constraint value can be chosen based on the required accuracy or based on the position of potential hazards along the flight path. The soft constraint is given by $|e_{p,xy}(t)| \leq e_{\text{max}} + \epsilon$, where $\epsilon$ is the slack control variable, which we add to the vector of control inputs:
\[
\bm{u} = \bigl[T \quad \phi_{\text{ref}} \quad \theta_{\text{ref}} \quad \epsilon \bigr]^T.
\]

\paragraph{Solver}
Solving the nonlinear OCP (\ref{eq:ocp}) at a high rate requires an efficient nonlinear solver routine. We formulate the MPC problem and generate a C++ interface using the ACADO toolkit for Matlab \cite{Houska2011a} and use a qpOASES solver \cite{Ferreau2014} with multiple-shooting discretization technique. The prediction horizon is set to 2 s, the discretization interval to 0.1 s.


\section{EXPERIMENTS}
\subsection{Implementation}
We implement the disturbance estimators and position controllers as C++ libraries wrapped in ROS \cite{Quigley2009} nodes. All flight tests are performed with a 700 g custom quadrotor based on an \textit{Asctec Hummingbird} frame equipped with a \textit{Pixhawk mini} autopilot connected to an \textit{Odroid XU4} onboard computer (2 GHz ARM processor, 2GB RAM). The state estimation and low-level controllers run on the \textit{Pixhawk} hardware, while the presented estimation and control algorithms run on two of the high-performance cores of the \textit{Odroid XU4}, leaving sufficient resources to run other elements of a typical UAS flight stack on the same processor.

\subsection{Testing Environments}
We first test and initially tune the C++ disturbance estimator and controller implementations in a \textit{RotorS}-based \cite{Furrer2016abc} simulation. We then fine-tune and validate the disturbance estimator in an indoor \textit{Vicon} environment where we use motion capture inputs for state estimation. Finally, we perform outdoor tests using an off-the-shelf GPS module for GPS-based state estimation. To ensure repeatibility and fair comparisons, the control experiments are conducted exclusively in the motion capture environment.

\subsection{Disturbance Estimation}
\label{subsec:disturbance_estimation}
\paragraph{Computational Cost}
We measure the iteration time of the EKF and UKF over the course of repeated landing maneuvers subject to wind gust disturbance. Both disturbance estimators are running in parallel on separate cores of the \textit{Odroid XU4} companion computer. The EKF iteration time averages at 8 ms, never exceeding 10 ms, whereas the UKF iteration time averages at around 20 ms, a 2.5-fold increase to the EKF. The additional computational cost of the UKF is largely due to the propagation of all sigma points and their recombination into a state mean and covariance at every timestep, involving a costly Cholesky matrix decomposition \cite{Tagliabue2017}. On the \textit{Odroid XU4}, the EKF can thus be run at a high update rate of 100 Hz, whereas the UKF can only be run at a maximum of 40 Hz.


\paragraph{Transient Performance}
\begin{figure}[t]
	\centering
	\begin{minipage}{.48\columnwidth}
		\centering
		\includegraphics[width=1.0\linewidth]{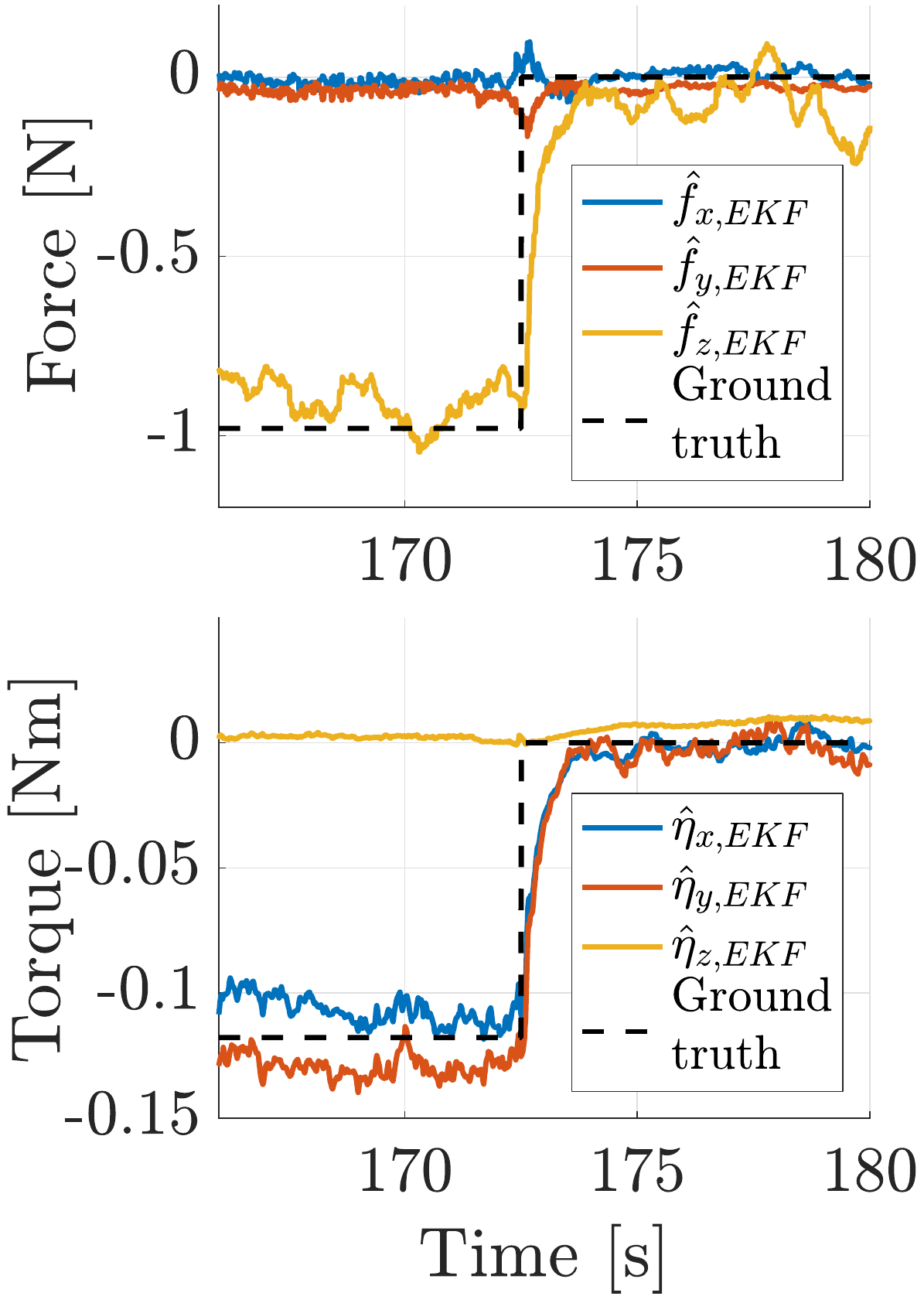}
		\caption{Step response of the EKF force and torque disturbance estimator, running at 100 Hz.}
		\label{fig:ekfstep}
	\end{minipage}%
	\hspace{1mm}
	\begin{minipage}{0.48\columnwidth}
		\vspace{2.5mm}
		\centering
		\includegraphics[width=1.0\linewidth]{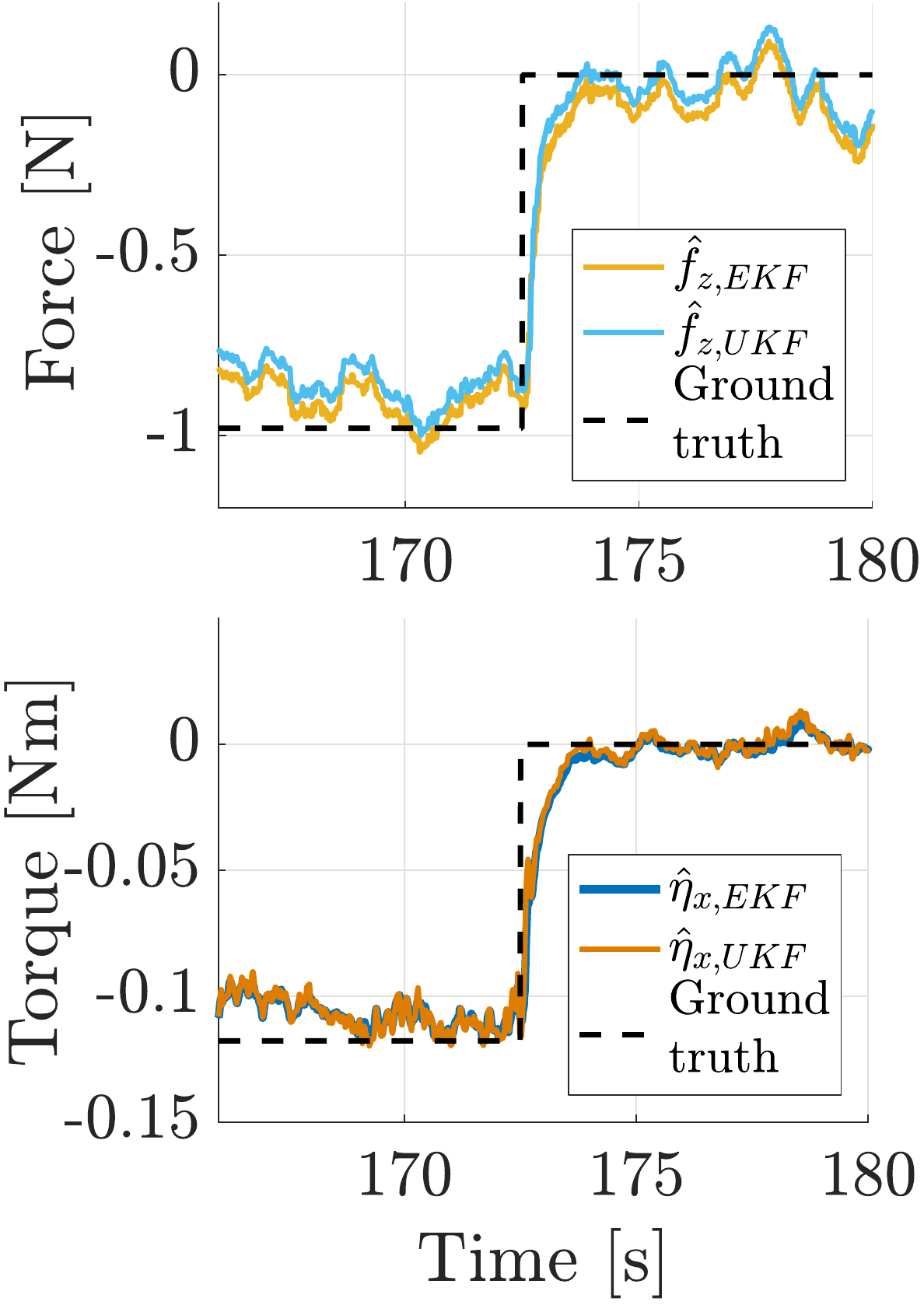}
		\caption{EKF/UKF comparison of the estimated disturbance force in z-axis and estimated roll torque disturbance.}
		\label{fig:ekfvsukf}
		\vspace{3mm}
	\end{minipage}
	\vspace{-7mm}
\end{figure}
\begin{figure}
	\vspace{4mm}
	\centering
	\includegraphics[width=1.0\linewidth]{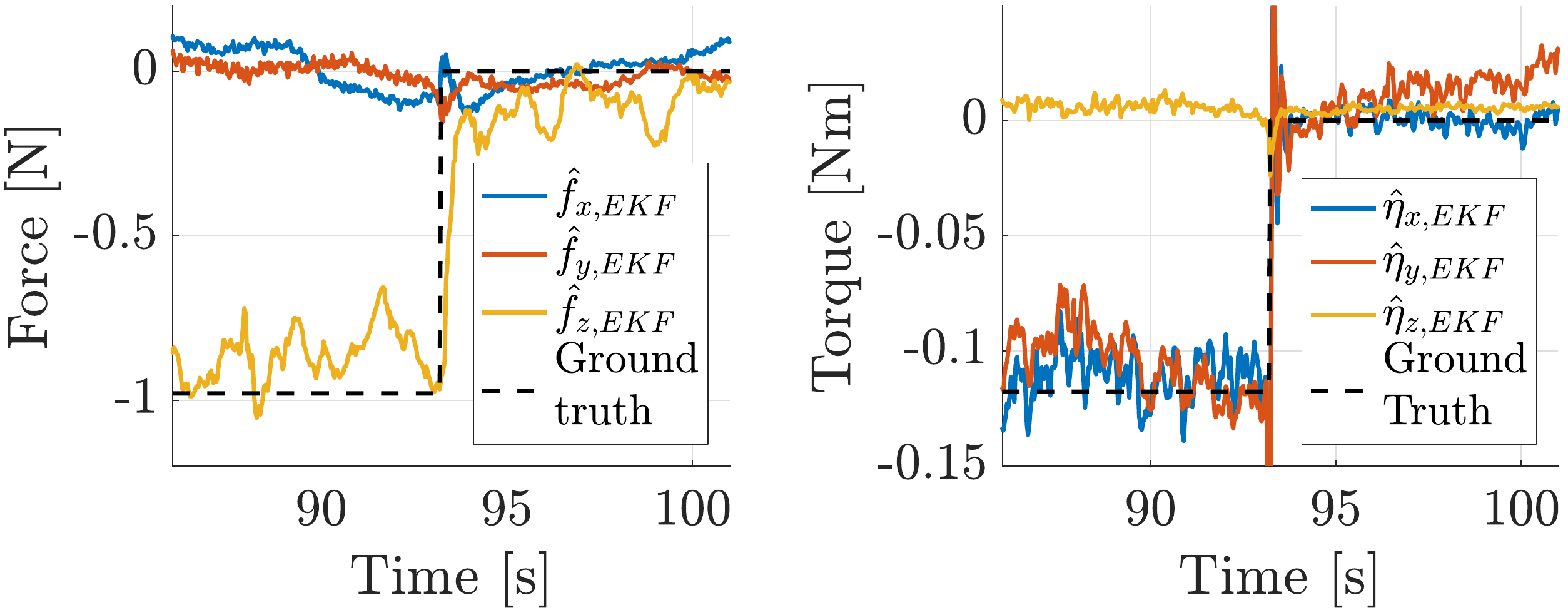}
	\caption{Step response of the EKF force and torque disturbance estimator outdoors using GPS-based state estimation.}
	\label{fig:gpsstep}
	\vspace{-3mm}
\end{figure}
To maximize rejection of fast disturbances such as wind gusts, a fast transient estimation is required. At the same time, overshoots and excessive noise in the disturbance estimate signals must be avoided to prevent a degradation of control performance. The tuning of the estimator covariance values is performed in a way that achieves this compromise. The transient performance is evaluated with a disturbance step response.
For this purpose we attach a weight ($m=0.1$ kg) to the upper right foot of the quadrotor, which we cut off in flight. Figure \ref{fig:ekfstep} shows that the EKF force and torque estimates converge to the ground truth values. The force estimate has a $t_{90}$ rise time of 1.1 s, while the torque estimate has a $t_{90}$ rise time of 1.0 s. The slight offset of the roll and pitch torque estimates to the ground truth can be explained by a slightly offset weight positioning.

To evaluate whether the UKF can improve on the EKF in terms of accuracy, convergence speed or noise, we run the UKF in parallel to the EKF at 40 Hz with the same variance settings. Figure \ref{fig:ekfvsukf} shows that the EKF and UKF estimates are nearly identical in terms of the above criteria. We conclude that the UKF does not offer superior performance and choose the EKF for disturbance estimation due to its reduced computational footprint.

To assess the impact of degraded state estimation on the performance of the EKF disturbance estimator, we repeat the experiment outdoors using GPS-based state estimation. Figure \ref{fig:gpsstep} shows that the disturbance estimator still provides accurate force and torque disturbance estimates. The force estimate exhibits similar convergence speed and only slightly increased noise compared to the estimate in Figure \ref{fig:ekfstep}. The torque estimate is considerably noisier, since the state estimator now relies purely on on-board sensing for attitude estimation, in contrast to the motion-capture-based state estimator used in the indoor experiment.

\subsection{Position Control - Computational Cost}

\begin{figure}[]
	\centering
	\includegraphics[width=0.9\linewidth]{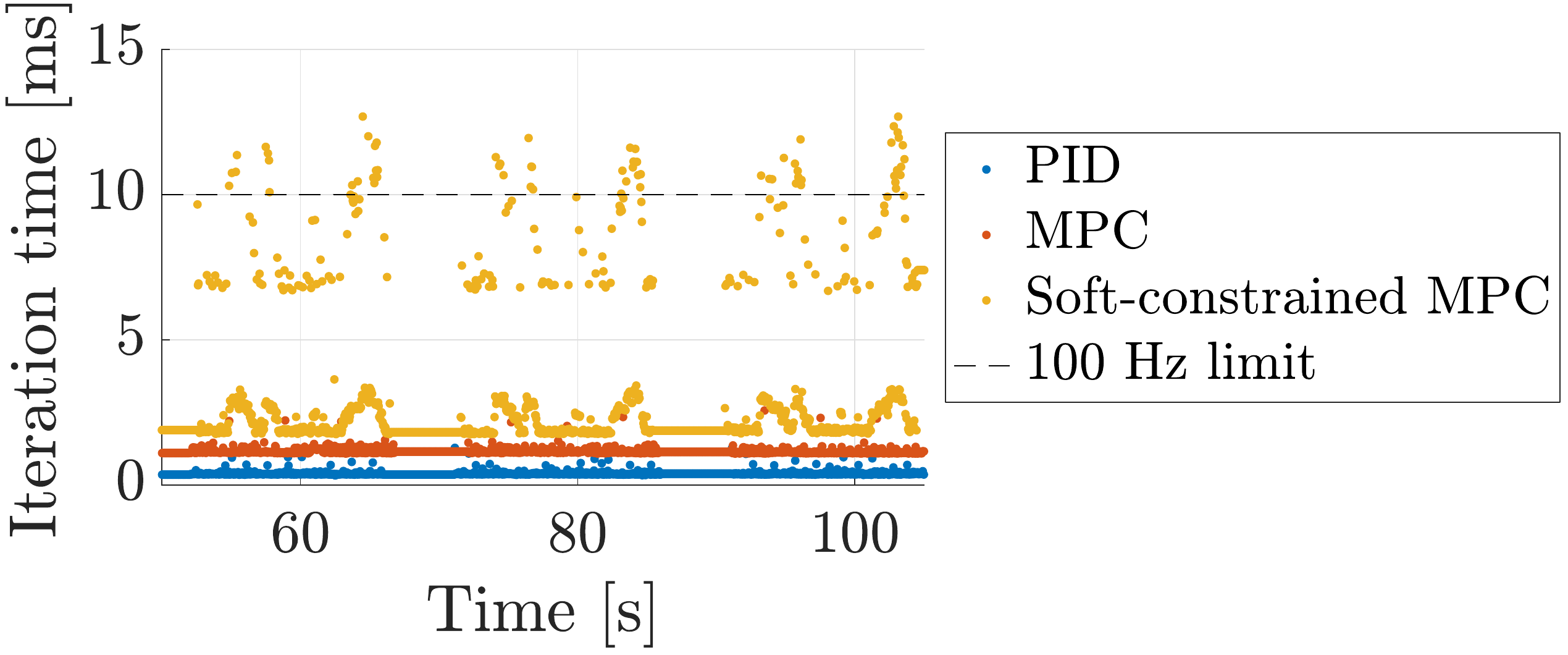}
	\caption{Comparison of iteration times between the PID, the MPC and the soft-constrained MPC path following controllers during repeated flights through a wind gust of 10 m/s.}
	\label{fig:controltimes}
	\vspace{-2mm}
\end{figure}

We measure the time it takes to run one control iteration over the course of repeated takeoff and landing maneuvers through a wind gust of around 10 m/s. Figure \ref{fig:controltimes} shows that the PID controller achieves iteration time average of 0.4 ms and maximum iteration times below 2 ms. The nonlinear MPC offers very efficient performance as well, with an iteration time average of 1.3 ms, with all measurements below 4 ms. Thus, the nonlinear MPC can easily be run at a fast update rate of 100 Hz on one core of the \textit{Odroid XU4}.

Computational cost increases significantly when adding an additional soft state constraint to the optimal control problem. In nominal conditions (no significant disturbances present), the average iteration time for the soft-constrained MPC is 2.5 ms. However, when subject to the strong wind gust disturbance, the original constraint is relaxed and the OCP takes significantly longer to solve, occasionally exceeding the 10 ms mark, but always staying below 13 ms. Still, the soft-constrained MPC can be run on the \textit{Odroid XU4} at a reasonably fast update rate of 70 Hz.
\label{subsec:cs_computational_cost}

\subsection{Ground Effect Disturbance Rejection}
\label{subsec:ground_effect}

\begin{figure}[]
	\vspace{4mm}
	\centering
	\begin{minipage}{1.0\columnwidth}
		\centering
		\includegraphics[width=1.0\linewidth]{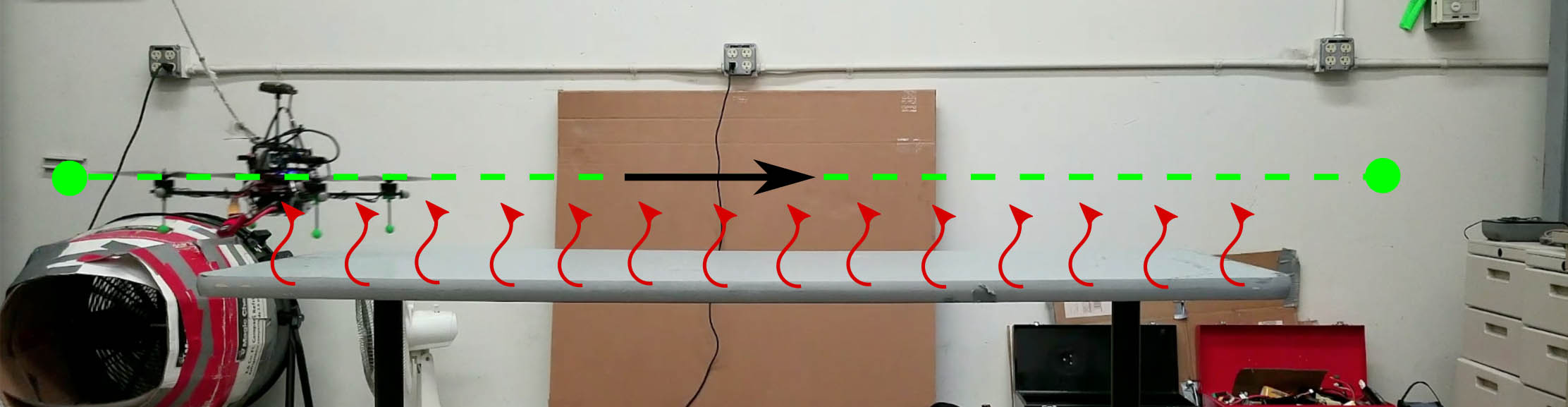}
		\caption{Ground effect experiment setup. Reference flight path in green. Red arrows represent the zone affected by ground effect.}
		\label{fig:groundeffectsetup}
		\vspace{-4mm}
	\end{minipage}%
	\vspace{6mm}
	\begin{minipage}{1.0\columnwidth}
		\vspace{4mm}
		\centering
		\includegraphics[width=1.0\linewidth]{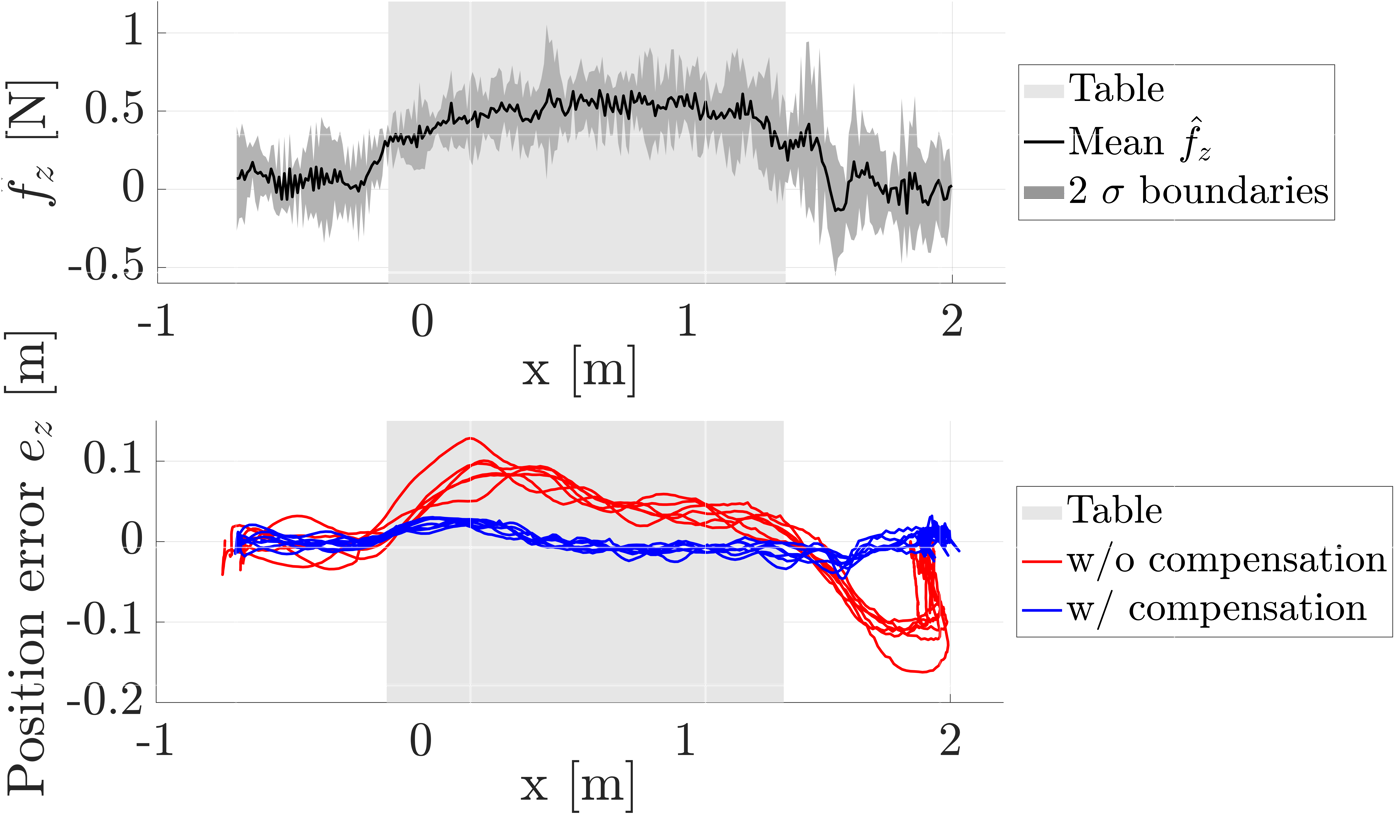}
		\caption{\textit{Top}: Mean and $2\sigma$ bounds of the estimated force disturbance along the z-axis during repeated flights over a table surface (see Figure 10). \textit{Bottom}: Position error along the z-axis for the flights with (in blue) and without (in red) disturbance rejection.}
		\label{fig:groundeffectcontrol}
		\vspace{-4mm}
	\end{minipage}
	\vspace{-2mm}
\end{figure}
\paragraph{Experiment Setup}
To test the ability of our framework to reject forces induced by ground effect, we perform repeated horizontal flights at a velocity of 0.3 m/s over the edges of a table, in close proximity (8 cm) to the table's surface. The experiment setup is illustrated in Figure \ref{fig:groundeffectsetup}. As we will show in \ref{subsec:single_gust}, there is little difference in the disturbance rejection capability for weak force disturbances between PID and MPC. Hence we perform this experiment using only the PID controller, with and without disturbance estimation and compensation. 

\paragraph{Results}
The top plot in Figure \ref{fig:groundeffectcontrol} shows that as soon as the vehicle crosses the left edge of the table, the disturbance force z-component converges to a positive value. As the vehicle crosses the opposing table edge, the disturbance estimate converges back to zero.
The bottom plot in Figure \ref{fig:groundeffectcontrol} shows that without disturbance rejection, the vehicle overshoots the position reference by 5 cm to 10 cm as it crosses the left table edge and enters the region affected by ground effect. As the vehicle flies over the surface, the integral error in the PID controller starts winding up, gradually correcting the error. However, as soon as the vehicle crosses the right edge of the table and exits the region affected by ground effect, the integral component of the controller causes the vehicle to undershoot the reference by around 10 cm. Only when the vehicle is commanded to hover at $x=2$ m does the integrator wind down, correcting the position error in the process.
With disturbance rejection, the disturbance feedforward term is able to compensate for the ground effect force disturbance, keeping the vehicle within 2.5 cm of the position reference at all times. No significant overshoots or undershoots are observed when the vehicle crosses the two table edges.

\subsection{Wind Gust Disturbance Rejection}
\label{subsec:single_gust}

\begin{figure}[t]
	\vspace{2mm}
	\centering
	\includegraphics[width=0.95\linewidth]{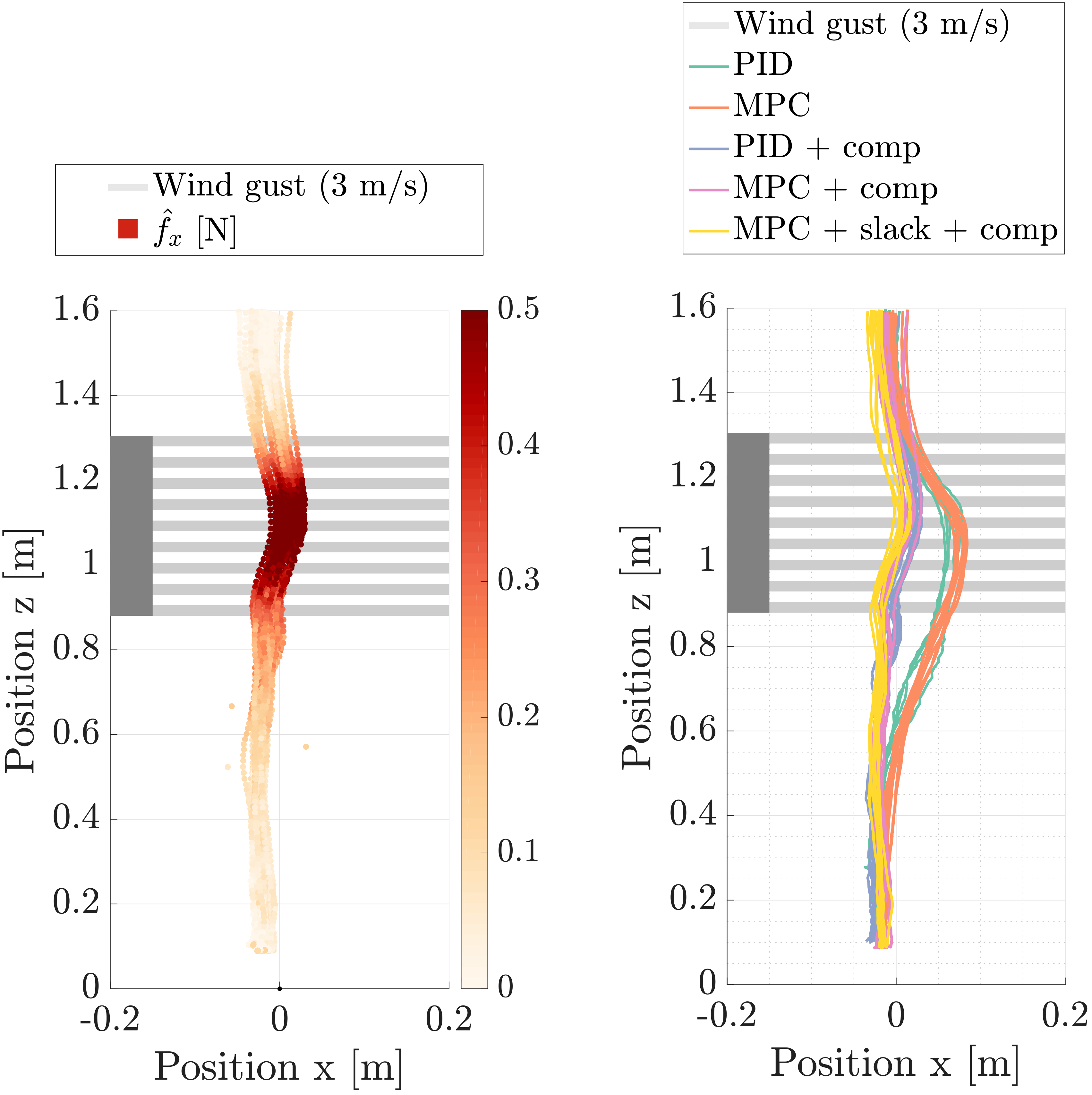}
	\caption{3 m/s wind gust experiment. \textit{Left}: Disturbance force estimate in Z-X plane during runs with disturbance rejection. \textit{Right}: Comparison of position error in Z-X plane during landing through strong wind gust for PID and MPC controllers, with and without disturbance rejection.}
	\label{fig:weakgust}
	\vspace{-8mm}
\end{figure}
\begin{figure}[t]
	\vspace{-1mm}
	\centering
	\includegraphics[width=0.88\linewidth]{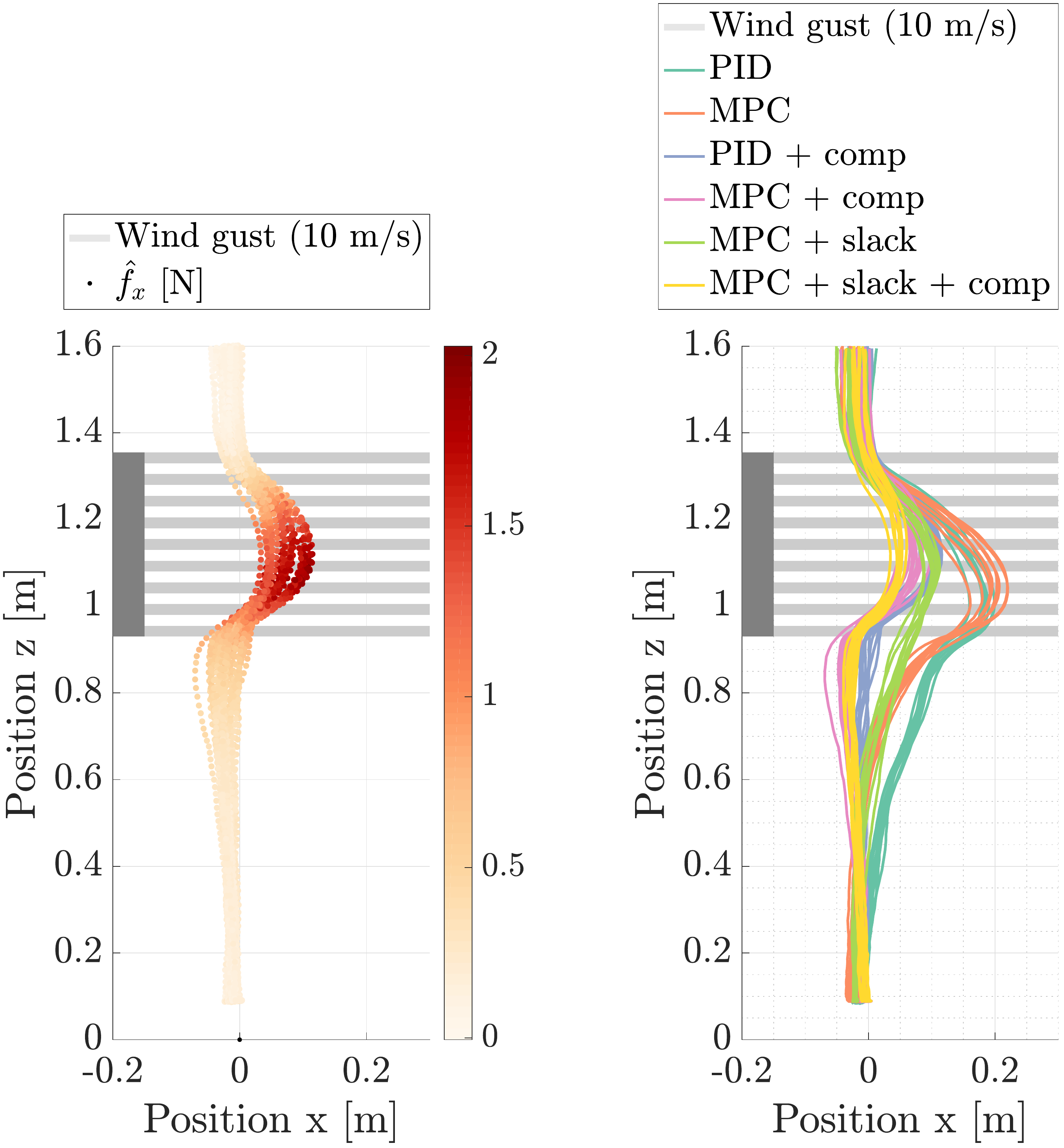}
	\caption{10 m/s wind gust experiment. \textit{Left}: Disturbance force estimate in Z-X plane during runs with disturbance rejection. \textit{Right}: Comparison of position error in Z-X plane during landing through strong wind gust for PID and MPC controllers, with and without disturbance rejection.}
	\label{fig:mediumgust}
	\vspace{-4mm}
\end{figure}
\begin{figure}[t]
	\vspace{-2mm}
	\centering
		\includegraphics[width=0.85\linewidth]{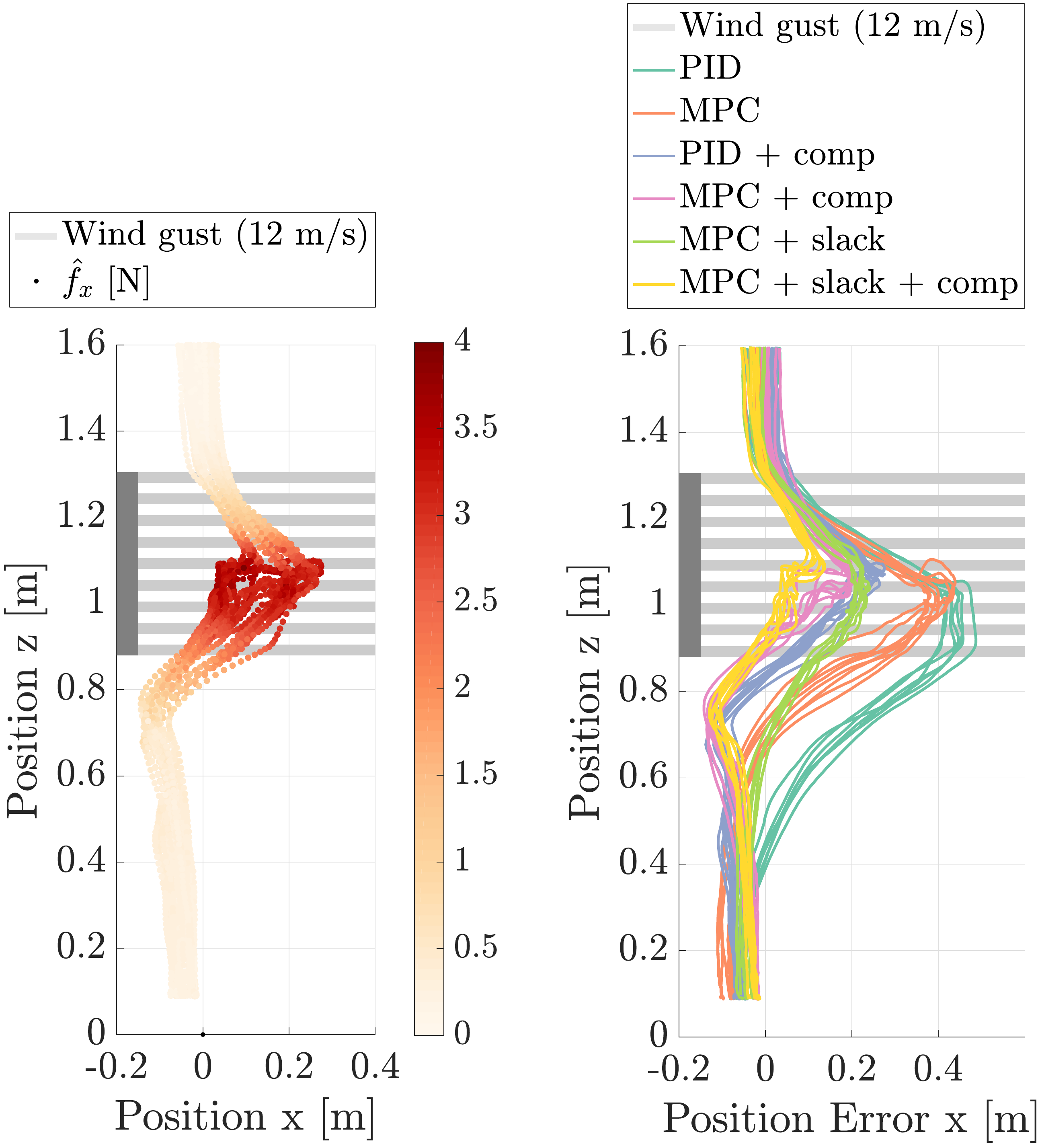}
	\caption{12 m/s wind gust experiment. \textit{Left}: Disturbance force estimate in Z-X plane during runs with disturbance rejection. \textit{Right}: Comparison of position error in Z-X plane during landing through strong wind gust for PID and MPC controllers, with and without disturbance rejection.}
	\label{fig:extremegust}
	\vspace{-4mm}
\end{figure}

\paragraph{Experiment Setup}
Next, we assess the robustness of the disturbance estimation and control system against strong, sudden wind gusts hitting the quadrotor in mid-air. For this purpose, we perform repeated vertical landings crossing a region affected by turbulent wind of different wind velocities produced by a high-powered fan. The wind velocities are measured using a digital anemometer in the center of the wind stream. The experiment setup is illustrated in Figure \ref{fig:setupgust}. We compare the disturbance rejection of the different control strategies by measuring the maximum deviations from the position reference. We also analyze the disturbance estimation performance. To enable a fair controller comparison, both the MPC and PID controllers were tuned to achieve an overdamped position step response with minimum rise time.

\begin{table}[b]
	\centering
	\caption{Comparison of maximum position error and improvement to baseline PID controller for strong wind gusts.}
	\label{tab:positionerror}
	\begin{tabular}{lcc}
		Wind velocity & 10 m/s& 12 m/s\\ \toprule
		Controller & \multicolumn{2}{c}{Max. horizontal deviation [cm]}\\ 
		\toprule
		PID (baseline)  & 17 & 45\\ 
		MPC  & 19 (-12\%) & 39 (+13\%)\\
		MPC + slack & 10 (+41\%) & 22 (+51\%) \\ \midrule
		PID + comp &  10 (+41\%)& 23 (+48\%)\\
		MPC + comp &  8 (+53\%)& 17 (+62\%) \\
		MPC + slack + comp &  5 (+71\%) & 10 (+78\%)\\
		\bottomrule
	\end{tabular}
\end{table}

\paragraph{Results}
Figure \ref{fig:weakgust} shows the disturbance estimation (left) and rejection performance (right) of the tested controllers for a wind velocity of 3 m/s, corresponding to conditions likely to be encountered outdoors on a slightly windy day. \textit{Without disturbance compensation}, both the PID and the MPC controllers experience an identical deviation from the reference path. \textit{With disturbance compensation}, the wind disturbance is almost completely rejected. No clear difference in performance between the controllers can be observed. 

To evaluate the estimation and control frameworks' performance in extreme conditions, we repeat the experiment with increased wind velocities. Figure \ref{fig:mediumgust} and \ref{fig:extremegust} show the same type of results for a wind velocity of 10 m/s and 12 m/s respectively, which could be encountered while flying in heavy winds or in narrow urban landscapes. Table \ref{tab:positionerror} summarizes the position errors and improvements to the baseline PID controller for the experiment with heavy winds. \textit{Without disturbance compensation}, MPC performs similarly to PID but features an improved recovery once the vehicle exits the wind field. Activating a slack constraint of 5 cm on the cross-error further improves the MPC performance, resulting in a 41\%-51\% improvement to the PID baseline. \textit{With disturbance compensation}, the disturbance rejection for both PID and MPC frameworks is significantly improved. However, the wind gust is not completely rejected. This is caused by the convergence time of the disturbance estimator. While increasing the variance of the additive Gaussian noise on the force estimate $\mathbf{v}_f$ decreases the convergence time, we found that the increased noise on the estimate translates to undesirable high-frequency roll and pitch oscillations. Hence, the requirement of smooth control imposes an upper bound on the magnitude of the additive noise variance, and thus the rejection performance. In both experiments, the best rejection performance is achieved by the slack-constrained MPC with disturbance compensation, featuring a considerable 71\%-78\% improvement over the baseline controller.

In the 12 m/s experiment, we observe that the MPCs with disturbance compensation control the vehicle to slow down the descent upon entering the wind gust. Due to the reduced penalty on the velocity error compared to the position error, the MPC prioritizes driving the position error to zero over maintaining a constant velocity. In case of very large disturbances, achieving both objectives is indeed not always possible due to motor saturation. This is a desirable property, since the vehicle first returns to its reference path before resuming forward flight.

\section{CONCLUSIONS}
In an extensive experimental comparison, we have shown that augmenting MPC multirotor position controllers with a disturbance estimator leads to significantly improved rejection of disturbances such as ground effect and fast wind gusts. We demonstrated that the developed MPC (in particular, the soft-constrained MPC) achieves higher levels of position accuracy in the presence of large external disturbances compared to a PID controller. Combined with its ability to run at a fast rate on a computationally restricted microprocessor, this makes MPC an attractive control strategy for applications with a need for precise position control despite dynamic external disturbances. Nonetheless, the developed PID framework offers much improved disturbance rejection compared to a conventional PID baseline.

For multirotor disturbance estimation, we have shown that an EKF framework is preferrable to an Unscented approach due to a 2.5-fold improved computational performance, reduced complexity and nearly identical estimation performance. We also demonstrated that the EKF disturbance estimator is suitable for deployment outdoors with degraded GPS-based state estimation.

An interesting direction for future work would be to combine model-based and data-driven approaches to learn to overcome disturbances, model inaccuracies or delays over time \cite{Pereida2017,Desaraju2017a}. Moreover, a direct comparison with robust and adaptive strategies would be desirable to benchmark position control accuracy subject to large external disturbances.




\section*{ACKNOWLEDGMENT}

This research was carried out at the Jet Propulsion Laboratory, California Institute of Technology, under a contract with the National Aeronautics and Space Administration.


\bibliographystyle{IEEEtranBST/IEEEtran}
\bibliography{bibliography_2/library}

\end{document}